\documentclass[12pt]{spieman}  
\usepackage{amsmath,amsfonts,amssymb}
\usepackage{todonotes,multirow}
\usepackage{graphicx}
\usepackage{setspace}
\usepackage{tocloft}
\usepackage{algorithm}
\usepackage{algorithmic}
\usepackage{color}
\usepackage{lineno}
\usepackage{multirow}
\usepackage[caption=false]{subfig}
\usepackage{textcomp} 
\usepackage{gensymb} 
\newcommand{\etal}{et al.~}
\newcommand{\ie}{i.e.}
\newcommand{\eg}{e.g.}

\title{Face recognition via compact second order image gradient orientations}

\author[a,b]{He-Feng Yin}
\author[a,b,*]{Xiao-Jun Wu}
\author[a,b]{Xiao-Ning Song}
\affil[a]{Jiangnan University, School of Artificial Intelligence and Computer Science, No. 1800, Lihu Avenue, Wuxi, China, 214122}
\affil[b]{Jiangsu Provincial Laboratory of Pattern Recognition and Computational Intelligence, No. 1800, Lihu Avenue, Wuxi, China, 214122}

\cftpagenumbersoff{figure}
\cftpagenumbersoff{table} 
\begin{document} 
\maketitle

\begin{abstract}
Conventional subspace learning approaches based on image gradient orientations only employ the first-order gradient information. However, recent researches on human vision system (HVS) uncover that the neural image is a landscape or a surface whose geometric properties can be captured through the second order gradient information. The second order image gradient orientations (SOIGO) can mitigate the adverse effect of noises in face images. To reduce the redundancy of SOIGO, we propose compact SOIGO (CSOIGO) by applying linear complex principal component analysis (PCA) in SOIGO. Combined with collaborative representation based classification (CRC) algorithm, the classification performance of CSOIGO is further enhanced. CSOIGO is evaluated under real-world disguise, synthesized occlusion and mixed variations. Experimental results indicate that the proposed method is superior to its competing approaches with few training samples, and even outperforms some prevailing deep neural network based approaches. The source code of CSOIGO is available at \url{https://github.com/yinhefeng/SOIGO}.
\end{abstract}


{\noindent \footnotesize\textbf{*}Xiao-Jun Wu,  \linkable{wu\_xiaojun@jiangnan.edu.cn} }

\begin{spacing}{2}   

\section{Introduction}
Face recognition (FR) remains one of the most active research topics in the community of pattern recognition, feature extraction \cite{zheng2006nearest,zheng2006reformative,wu2004new} is a key ingredient in FR, as well as in image fusion \cite{luo2016novel,luo2017image,li2017multi,li2020nestfuse} and many other computer vision tasks\cite{li2011no,chen2018new,sun2019effective,wang2003initial,sun2011quantum}. Though considerable progress has been made during the past decades, robust FR is still unresolved. Occlusion is ubiquitous in practical applications, and it will dramatically degrade the performance of FR. To increase the robustness to occlusion, researchers have developed a variety of approaches. Sparse representation based classification (SRC) \cite{wright2008robust} is developed for FR and it shows robustness to occlusion and corruption in the test images when combined with block partition technique. Naseem \etal \cite{naseem2010linear} proposed a modular linear regression classification (Modular LRC)  approach with a distance based evidence fusion (DEF) algorithm to tackle the problem of contiguous occlusion. To further enhance the performance of SRC, Li \etal \cite{li2019sparsity} proposed a sparsity augmented weighted CRC approach for image recognition. Dong \etal \cite{dong2019low} designed a low-rank Laplacian-uniform mixed (LR-LUM) model which characterizes complex errors as a combination of continuous structured noises and random noises. Yang \etal \cite{yang2017nuclear} presented nuclear norm based matrix regression (NMR) which employs two dimensional image-matrix-based error model rather than the one dimensional pixel-based error model. The representation vector in NMR is imposed by the $\ell_2$ norm, to make use of the discriminative property of sparsity, Chen \etal \cite{chen2019sparse} proposed a sparse regularized NMR (SR-NMR) by replacing the $\ell_2$ norm constraint on the representation vector with the $\ell_1$ norm. However, the above approaches need uncorrupted training images. When providing corrupted training data, their performance will be deteriorated. To tackle the situation that both the training and test data are corrupted, low rank matrix recovery (LRMR) can be applied. Chen \etal \cite{chen2014sparse} proposed a discriminative low rank representation (DLRR) method which introduces the structural incoherence into the framework of low rank representation (LRR) \cite{liu2012robust}. Gao \etal \cite{gao2017learning} proposed to learn robust and discriminative low-rank representation (RDLRR) by introducing low-rank constraint to simultaneously model the representation and each error term. Hu \etal \cite{hu2018robust} presented a robust FR method which employs dual nuclear norm low rank representation and self-representation induced classifier. Yang \etal \cite{yang2018sparse} developed a sparse low-rank component-based representation (SLCR) method for FR with low quality images. Recently, Yang \etal \cite{yang2021sparse} extended SLCR and proposed a FR technique named sparse individual low-rank component representation (SILR) for IoT-based system. Inspired by LRR and deep learning techniques, Xia \etal \cite{xia2020embedded} developed an embedded conformal deep low-rank auto-encoder (ECLAE) neural network architecture for matrix recovery.

Recently, image gradient orientations (IGO) has attracted much attention due to its impressive results in occluded FR. Wu \etal \cite{wu2018occluded} presented a gradient direction-based hierarchical adaptive sparse and low rank (GD-HASLR) model which performs in the image gradient direction domain rather than the image intensity domain. Li \etal \cite{li2019image} incorporated IGO into robust error coding and proposed an IGO-embedded structural error coding (IGO-SEC) model for FR with occlusion. Apart from the above two works, Zhang \etal \cite{zhang2009face} designed Gradientfaces for FR under varying illumination conditions. In essence, Gradientfaces is the IGO. Tzimiropoulos \etal \cite{tzimiropoulos2012subspace} introduced the notion of subspace learning from IGO and developed approaches such as IGO-PCA and IGO-LDA. Vu \cite{vu2012exploring} proposed a face representation approach called patterns of orientation difference (POD) which explores the relations of both gradient orientations and magnitudes. Zheng \etal \cite{zheng2019online} presented an online image alignment method via subspace learning from IGO. Qian \etal \cite{qian2020image} presented a method called ID-NMR in which the local gradient distribution is exploited to decompose the image into several gradient images.

The above IGO based approaches only take the first order gradient information into account, thus neglecting the second order or higher order gradient information. Latest researches on human vision discover that the neural image is a landscape or a surface whose geometric properties can be described by local curvatures of differential geometry through second order gradient information \cite{huang2014hsog,morgan2011features}. Based on the second order gradient, Huang \etal \cite{huang2014hsog} presented a new local image descriptor called histograms of second order gradient (HSOG). Li \etal \cite{li2019defect} proposed a patterned fabric defect detection method based on the second-order orientation-aware descriptor. Zhang \etal \cite{zhang2020no} designed a blind image quality assessment (IQA) method based on multi-order gradients statistics. Bastian \etal \cite{bastian2019pedestrian} developed a pedestrian detector utilizing both the first order and the second order gradient information in the image. Nevertheless, the above second order gradient based approaches do not involve dimensionality reduction technique which would result in redundant information. To alleviate this problem, we introduce PCA into the framework of SOIGO to extract more compact features. Moreover, we employ CRC as the final classifier due to its effectiveness and efficiency. Experimental results show that our proposed method (CSOIGO) is robust to real disguise, synthesized occlusion and mixed variations, and it is superior to some popular deep neural network based approaches.

The remainder of this paper is arranged as follows. Section \ref{sec_2} reviews some related work. In Section \ref{sec_3}, we present our proposed approach. Section \ref{sec_4} conducts several experiments to demonstrate the efficacy of our proposed method. Finally, conclusions are drawn in Section \ref{sec_5}.
\section{Related work}
\label{sec_2}
\subsection{IGO-PCA}
Given a set of images $\left \{ \mathbf{Z}_i \right \}$ ($i=1,2,\ldots,N$), where $N$ denotes the number of training images and $\mathbf{Z}_i \in \mathbb{R}^{m \times n}$. Suppose $\mathbf{I}(x,y)$ is the image intensities at pixel coordinates $(x,y)$ of sample $\mathbf{Z}_i$, the horizontal and vertical gradient can be obtained by the following formulations:
\begin{equation}
\begin{aligned}
&\mathbf{G}_{i,x}=h_x*\mathbf{I}(x,y) \\
&\mathbf{G}_{i,y}=h_y*\mathbf{I}(x,y)
\end{aligned}
\end{equation}
where $*$ expresses convolution, $h_x$ and $h_y$ are filters employed to approximate the ideal differentiation operator along the image horizontal and vertical directions, respectively. However, the image data mostly distribute discretely in real-world scenarios, so we usually use differences to compute the gradients, \ie, achieving the gradients through the difference between adjacent pixels' gray values. Thus horizontal and vertical gradients can be reformulated as:
\begin{equation}
\label{eq:first_grad}
\begin{aligned}
&\mathbf{G}_{i,x}=\mathbf{I}(x+1,y)-\mathbf{I}(x,y) \\
&\mathbf{G}_{i,y}=\mathbf{I}(x,y+1)-\mathbf{I}(x,y)
\end{aligned}
\end{equation}
Then the gradient orientation of the pixel location $(x,y)$ is:
\begin{equation}
\label{eq:gradori}
\Phi_i(x,y)=\textrm{arctan}\tfrac{\mathbf{G}_{i,y}}{\mathbf{G}_{i,x}},i=1,2,...,N 
\end{equation}
For each image $\mathbf{Z}_i$ whose size is $m \times n$, we can get a corresponding gradient orientation matrix $\Phi _i\in [0,2\pi)^{m\times n}$. Then we can obtain the corresponding sample vectors by converting 2D images $\Phi_i$ into 1D vectors $\mathbf{\phi} _i$. Referring to Ref.\citenum{tzimiropoulos2012subspace}, we also define the mapping from $[0,2\pi)^K(K=m \times n)$ onto a subset of complex sphere with radius $\sqrt{K}$
\begin{equation}
\label{eq:mapping}
\boldsymbol{t}_i(\phi_i)=e^{j\phi_i}
\end{equation}
where $e^{j\phi_i}=[e^{j\phi_1},e^{j\phi_2},...,e^{j\phi_K}]^T $, and  $e^{j\theta } $ is Euler form,  \ie, $e^{j\theta}=\textrm{cos}\theta+j\textrm{sin}\theta $.
Then we can apply complex linear PCA to the transformed $\boldsymbol{t}_i$. That is, we seek for a set of $d<K$ orthonormal bases $\mathbf{U}=[\mathbf{u}_1,\mathbf{u}_2,...,\mathbf{u}_d]\in \mathbb{C}^{K\times d}$ by solving the following problem:
\begin{equation}
\label{eq:pca_obj}
\epsilon (\mathbf{U})=\left \| \mathbf{X}-\mathbf{U}\mathbf{U}^{H}\mathbf{X} \right \|_F^2
\end{equation}
where $\mathbf{X}=[\boldsymbol{t}_1,\boldsymbol{t}_2,...,\boldsymbol{t}_N]\in \mathbb{C}^{K\times N} $, $\mathbf{U}^H$ is the conjugate transpose of $\mathbf{U}$, and $\left \| . \right \|_F$ denotes the Frobenius norm. Eq. \ref{eq:pca_obj} can be reformulated as:
\begin{equation}
\label{eq:equ_obj}
\mathbf{U}_o=\textrm{arg} \ \underset{\mathbf{U}}{\textrm{max}} \ tr(\mathbf{U}^{H}\mathbf{X}\mathbf{X}^{H}\mathbf{U}), \ \textrm{s.t.} \ \mathbf{U}^{H}\mathbf{U}=\mathbf{I} 
\end{equation}
The solution is given by the $d$ eigenvectors of $\mathbf{X}\mathbf{X}^H$ corresponding to the $d$ largest eigenvalues. Then the $d$-dimensional embedding $\mathbf{Y}\in \mathbb{C}^{d\times N}$ of $\mathbf{X}$ is produced by $\mathbf{Y}=\mathbf{U}^H\mathbf{X}$.

\subsection{Collaborative representation based classification}
During the past few years, representation based classification method (RBCM) has attracted lots of attention in the community of pattern recognition. The pioneering work is SRC \cite{wright2008robust}. In SRC, the $\ell_1$ norm constraint is employed to attain the sparse coefficient of test data. Afterwards, Zhang \etal \cite{zhang2011sparse} argued that it is the collaborative representation mechanism rather than the $\ell_1$ norm constraint that makes SRC successful for FR. Therefore, they developed the CRC method which replaces the $\ell_1$ norm constraint with the $\ell_2$ norm, and the objective function of CRC is formulated as follows,
\begin{equation}
\label{eq:crc_obj}
\underset{\boldsymbol{\alpha}}{\textrm{min}} \ \left \{ \left \| \boldsymbol{y}-\mathbf{D}\boldsymbol{\alpha} \right \|_2^2+\lambda\left \| \boldsymbol{\alpha} \right \|_2^2 \right \}
\end{equation}
where $\boldsymbol{y}$ is the test data, $\mathbf{D}$ is the dictionary that contains all the training data from $C$ classes, and $\lambda$ is a balancing parameter. Eq. \ref{eq:crc_obj} has the following closed-form solution,
\begin{equation}
\label{eq:crc_solution}
\boldsymbol{\alpha}=(\mathbf{D}^T\mathbf{D}+\lambda \mathbf{I})^{-1}\mathbf{D}^T\boldsymbol{y}
\end{equation}
In the classification stage, apart from the class specific reconstruction error $\left \| \boldsymbol{y}-\mathbf{D}_j\boldsymbol{\alpha}_j \right \|_2,j=1,2,\ldots,C$, where $\boldsymbol{\alpha}_j$ is the coefficient vector corresponding to the $j$th class, Zhang \etal \cite{zhang2011sparse} found that $\left \| \boldsymbol{\alpha}_j \right \|_2$ also contains some discriminative information for classification. Thus, they presented the following regularized residuals for classification,
\begin{equation}
\label{eq:crc_rule}
\textrm{identity}(\boldsymbol{y})=\textrm{arg} \ \underset{j}{\textrm{min}} \ \frac{\left \| \boldsymbol{y}-\mathbf{D}_j\boldsymbol{\alpha}_j \right \|_2}{\left \| \boldsymbol{\alpha}_j \right \|_2}
\end{equation}

\section{Proposed method}
\label{sec_3}
\begin{figure}[t]
\centering
\includegraphics[trim={0mm 0mm 0mm 0mm},clip, width = .8\textwidth]{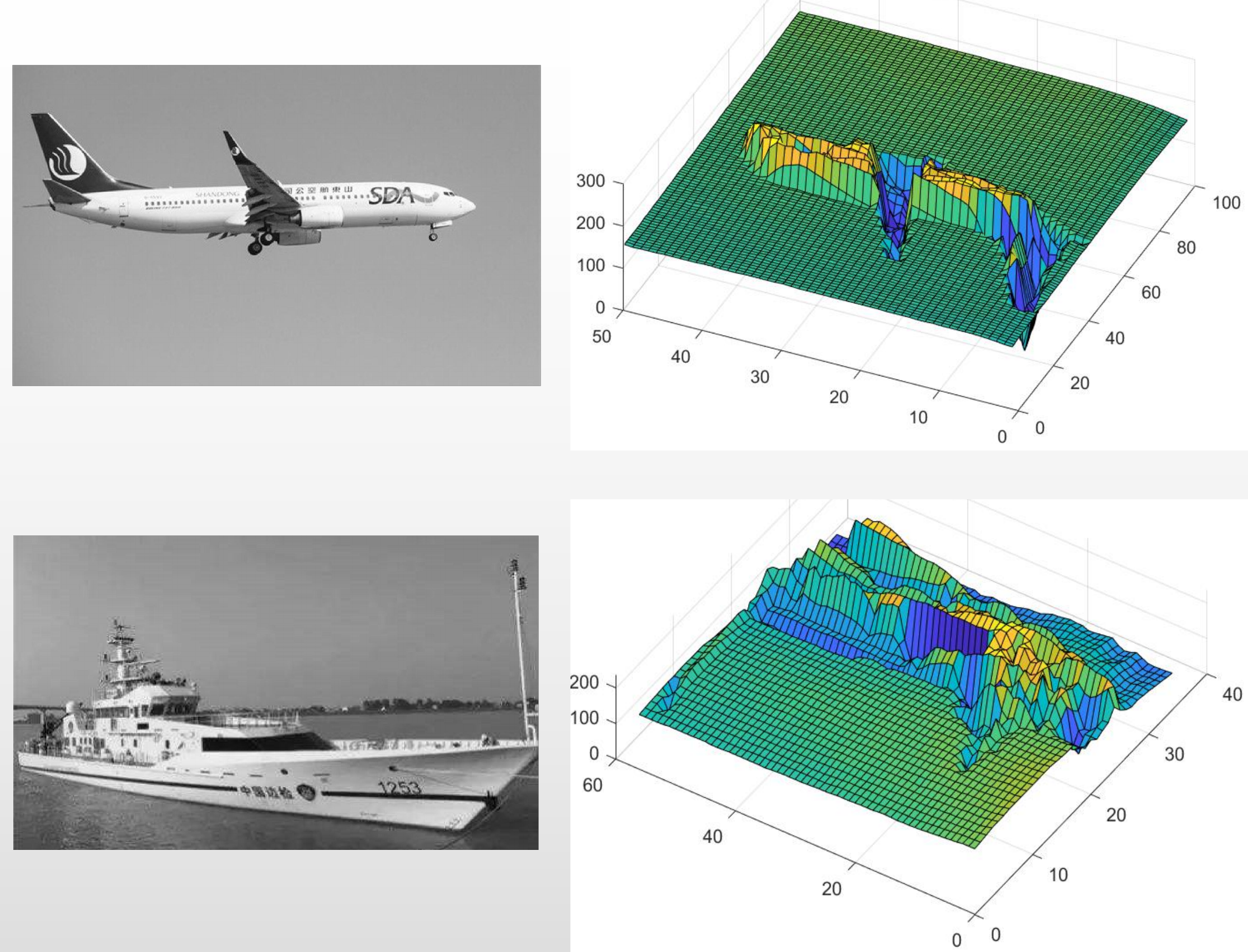}
\caption{Original images (left part) and their surface plots (right part).}
\label{fig:landscape}
\end{figure}

Previous studies reveal that gradient information at different orders characterize different structural features of natural scenes. The first-order gradient information is related to the slope and elasticity of a surface, while the second order gradient delivers the curvature related geometric properties. Fig. \ref{fig:landscape} depicts two images and their corresponding landscapes plotted as surfaces, one can see that these landscapes contain a variety of local shapes,  such as cliffs, ridges, summits, valleys and basins. Inspired by the above results, we propose a new FR method which exploits the SOIGO. And the second order gradient is obtained based on the first order gradient information defined in Eq. \ref{eq:first_grad},

\begin{equation}
\label{eq:second_grad}
\begin{aligned}
&\mathbf{G}_{i,x}^{2}=\mathbf{G}_{i,x}(x+1,y)-\mathbf{G}_{i,x}(x,y) \\
&\mathbf{G}_{i,y}^{2}=\mathbf{G}_{i,y}(x,y+1)-\mathbf{G}_{i,y}(x,y)
\end{aligned}
\end{equation}
where $\mathbf{G}_{i,x}^{2}$ and $\mathbf{G}_{i,y}^{2}$ are the second order gradient along the horizontal and vertical directions, respectively. Therefore, the SOIGO is computed as follows,
\begin{equation}
\label{eq:soigo}
\Phi_i^{2}(x,y)=\textrm{arctan}\tfrac{\mathbf{G}_{i,y}^{2}}{\mathbf{G}_{i,x}^{2}}
\end{equation}

\begin{figure}[t]
\centering
\includegraphics[trim={0mm 0mm 0mm 0mm},clip, width = .8\textwidth]{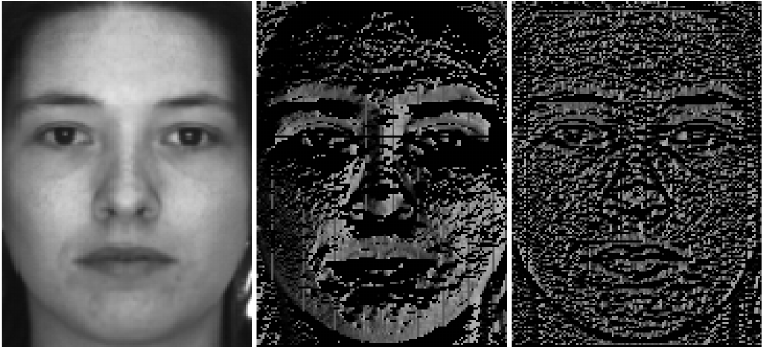}
\caption{Original face image and its gradient orientations of the first and second orders, respectively.}
\label{fig:IGO_img}
\end{figure}

\begin{figure}[t]
    \centering
    \subfloat[original data]{
    \includegraphics[trim={0mm 0mm 0mm 0mm},clip, width =.3\textwidth]{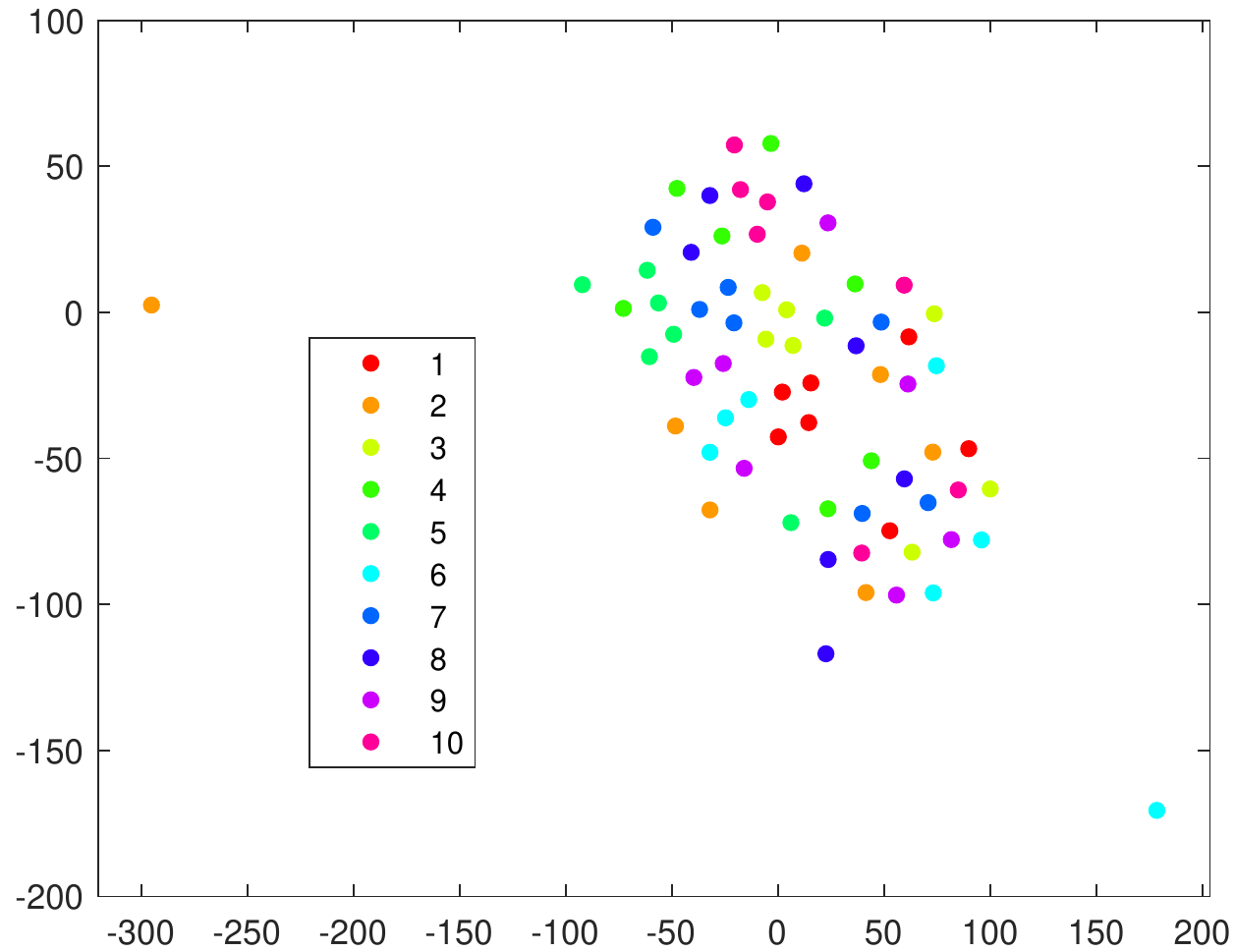}
    \label{fig:tsne_ori}
    }
    \subfloat[the first order IGO]{
    \includegraphics[trim={0mm 0mm 0mm 0mm},clip, width =.3\textwidth]{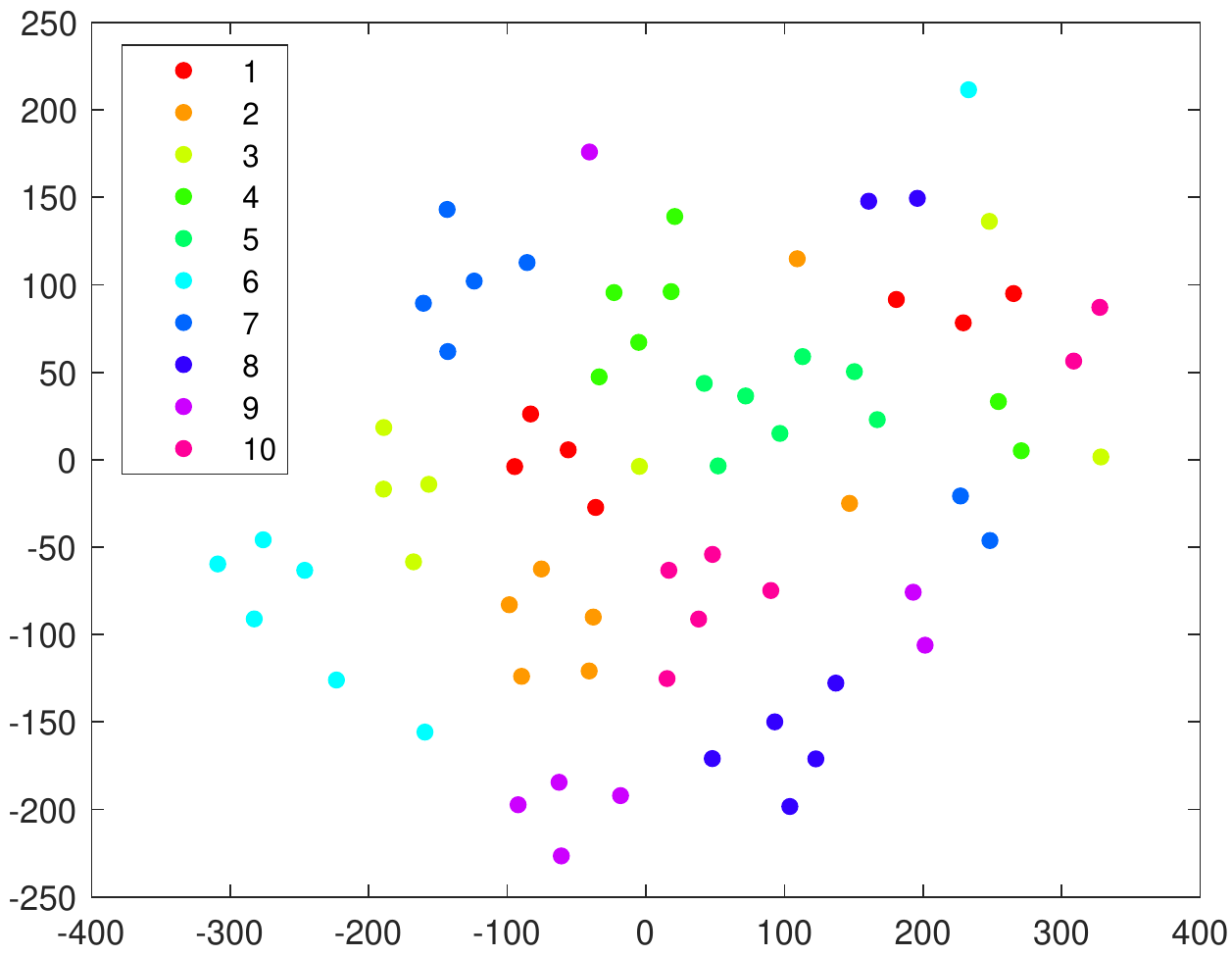}
    \label{fig:tsne_igo}
    }
\subfloat[the SOIGO]{
    \includegraphics[trim={0mm 0mm 0mm 0mm},clip, width =.3\textwidth]{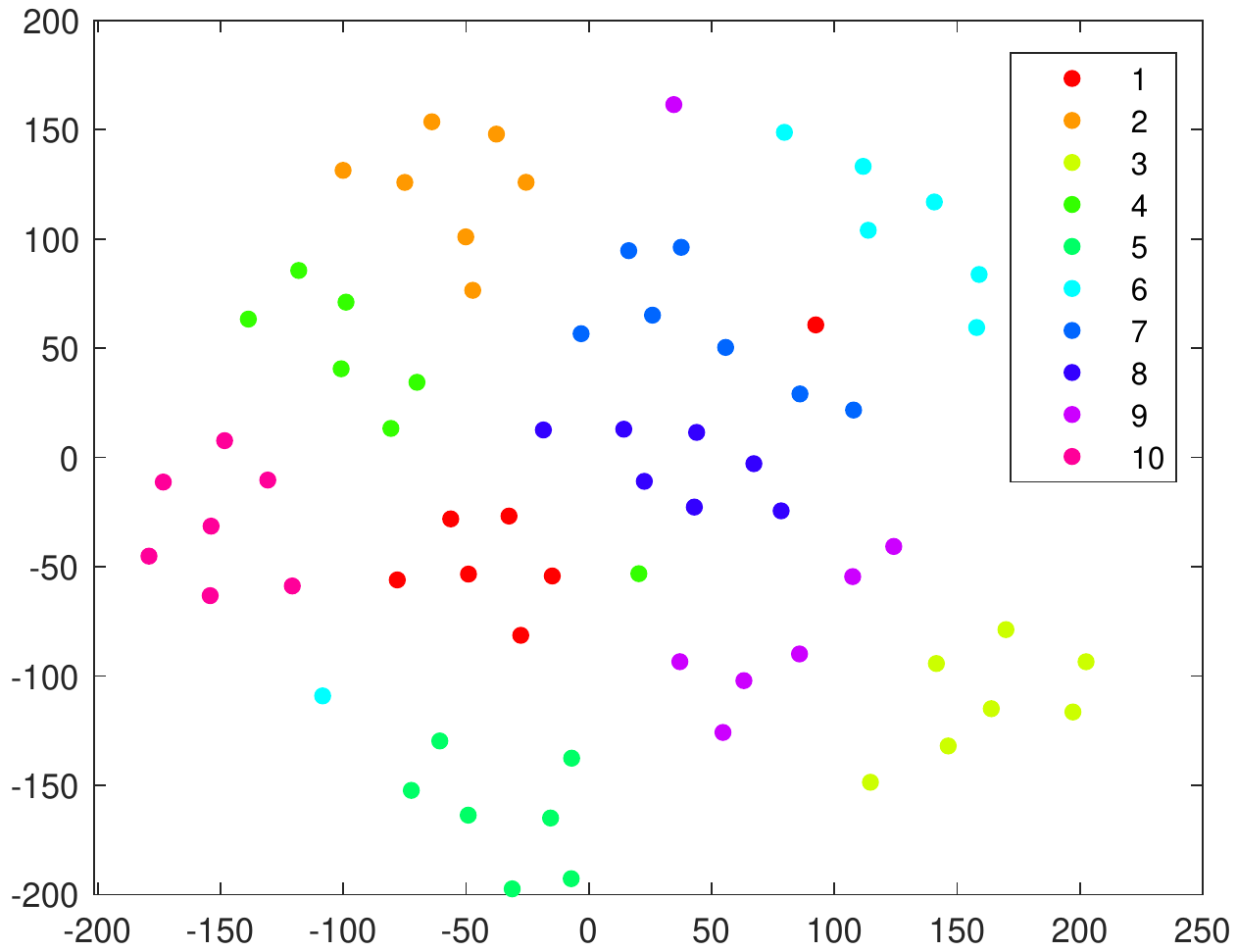}
    \label{fig:tsne_soigo}
    }
\caption{t-SNE visualization of (a) original data, (b) the first order IGO with the mapping defined in Eq. \ref{eq:mapping} and (c) the SOIGO with the mapping defined in Eq. \ref{eq:mapping}. For better visualization, please refer to the electronic version of this paper.}
\label{fig:tsne}
\end{figure}
Fig. \ref{fig:IGO_img} presents an original face image and its gradient orientations of the first and second orders, one can see that, compared with the first order IGO, the SOIGO significantly depress the noises in the orientation domain. Moreover, the SOIGO contains more fine information than the first order IGO, \eg, areas around the eyes, nose and mouth.

To further illustrate the effectiveness of using the SOIGO, we visualize the original data, the first order IGO and the SOIGO on the AR database by employing the t-SNE algorithm \cite{maaten2008visualizing} in Fig. \ref{fig:tsne}. These data are selected from the first ten subjects on the AR database, for each person, seven non-occluded face images in Session 1 are used. Then these images are occluded by a square baboon image with a percentage of 30\%, for detailed experimental settings, please refer to subsection \ref{sec_43}. As can be seen from Fig. \ref{fig:tsne}, though the first order IGO looks better as compared with the original data, clusters of different classes are mixed together. In Fig. \ref{fig:tsne} (c), cluster of the same class is more compact than that of Fig. \ref{fig:tsne} (b), which is beneficial for subsequent classification.

The procedures of obtaining the projection matrix $\mathbf{U}$ is the same as in IGO-PCA. Then for a test image $\mathbf{Z}_t$, first we compute its SOIGO and obtain $\boldsymbol{t}$ after the mapping defined by Eq. \ref{eq:mapping}.
Embeddings of training and test images are derived as follows,
\begin{equation}
\label{eq:tr_embedding}
\mathbf{Y}=\mathbf{U}^H\mathbf{X}
\end{equation}

\begin{equation}
\label{eq:tt_embedding}
\boldsymbol{z}=\mathbf{U}^H\boldsymbol{t}
\end{equation}
where $\mathbf{Y} \in \mathbb{C}^{d\times N}$ and $\boldsymbol{z} \in \mathbb{C}^{d\times 1}$. To make the embeddings of training and test images suitable for CRC, we employ both the real and imaginary parts of $\mathbf{Y}$ and $\boldsymbol{z}$ as the input of CRC, let
\begin{equation}
\label{eq:dict_trans}
\mathbf{D}=\begin{bmatrix}
\textrm{real}(\mathbf{Y}) \\ 
\textrm{imag}(\mathbf{Y}) 
\end{bmatrix}
\end{equation}

\begin{equation}
\label{eq:test_trans}
\boldsymbol{y}=\begin{bmatrix}
\textrm{real}(\boldsymbol{z}) \\ 
\textrm{imag}(\boldsymbol{z}) 
\end{bmatrix}
\end{equation}
where $\textrm{real}(\cdot)$ and $\textrm{imag}(\cdot)$ are the real part and imaginary part of complex number, respectively. Then we compute the representation coefficient vector of $\boldsymbol{y}$ over $\mathbf{D}$, this is followed by checking which class results in the least regularized residual. The complete process of our proposed CSOIGO is outlined in Algorithm \ref{alg1}.
\begin{algorithm}[t]
\begin{algorithmic}
\vspace{0.03in}
\STATE \textbf{Input:} A set of $N$ training images $\left \{ \mathbf{Z}_i \right \}(i=1,2,\ldots,N)$ from $C$ classes, test image $\mathbf{Z}_t$, the number of principal components $d$ and the regularization parameter $\lambda$ for CRC.

\STATE \ \ \ \ 1. Obtain the SOIGO $\Phi_i^2$ of training images and convert it to 1D vector $\phi_i^2$.
\STATE \ \ \ \ 2. Compute $\mathbf{t}_i(\phi_i^2)=e^{j\phi_i^2}$, all the SOIGO of training images form the matrix $\mathbf{X}=[\mathbf{t}_1,\mathbf{t}_2,...,\mathbf{t}_N]$.
\STATE \ \ \ \ 3. Obtain the projection matrix $\mathbf{U}$ via Eq. \ref{eq:equ_obj}.
\STATE \ \ \ \ 4. For the test image $\mathbf{Z}_t$, obtain its SOIGO $\Phi_t^2$ and convert it to 1D vector $\phi_t^2$, then compute $\boldsymbol{t}=e^{j\phi_t^2}$.
\STATE \ \ \ \ 5. Obtain the embeddings of training and test images via Eqs. \ref{eq:tr_embedding} and \ref{eq:tt_embedding}.
\STATE \ \ \ \ 6. Obtain $\mathbf{D}$ and $\boldsymbol{y}$ by Eqs. \ref{eq:dict_trans} and \ref{eq:test_trans}.
\STATE \ \ \ \ 7. Code $\boldsymbol{y}$ over $\mathbf{D}$ by Eq. \ref{eq:crc_solution}.
\STATE \ \ \ \ 8. Compute the regularized residuals $\boldsymbol{r}_j=\frac{\left \| \boldsymbol{y}-\mathbf{D}_j\boldsymbol{\alpha}_j \right \|_2}{\left \| \boldsymbol{\alpha}_j \right \|_2},j=1,2,\ldots,C$.
\STATE \textbf{Output:} $\textrm{identity}(\mathbf{Z}_t)=\textrm{arg} \ \underset{j}{\textrm{min}} \ \boldsymbol{r}_j$.
\vspace{0.03in}
\end{algorithmic}
\caption{CSOIGO}
\label{alg1}
\end{algorithm}

\section{Experimental results and analysis}
\label{sec_4}
In this section, experiments are conducted under different scenarios to validate the effectiveness of the proposed method.
\subsection{Recognition with real disguise}
\label{sec_41}
\begin{figure}[t]
\centering
\includegraphics[trim={0mm 0mm 0mm 0mm},clip, width = .8\textwidth]{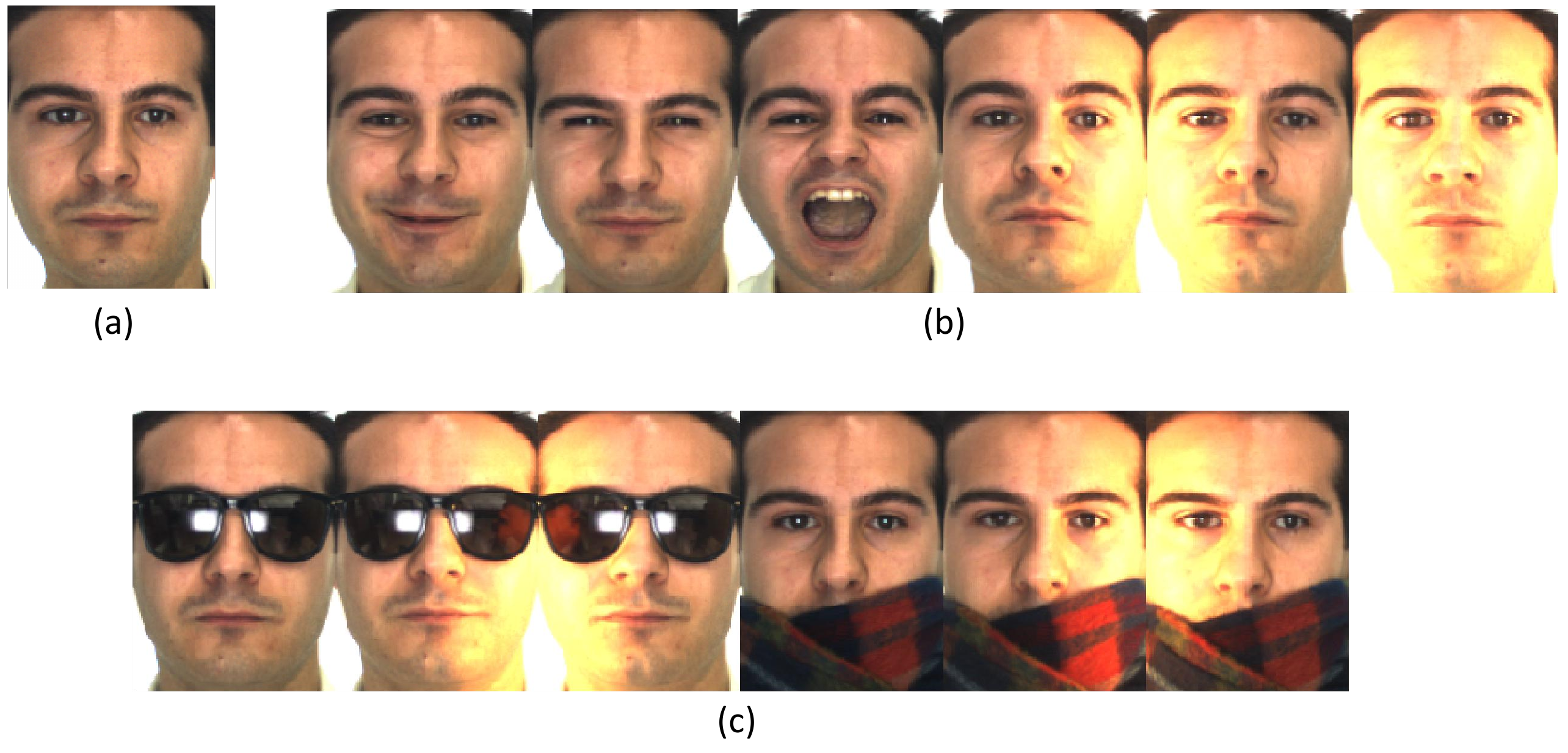}
\caption{Some example face images from the AR database, (a) the neutral image of a subject from Session 1; (b) face images with illumination and expression variations; (c) occluded images by sunglasses/scarf.}
\label{fig:AR_img}
\end{figure}

The AR database contains over 4000 images of 126 subjects. For each individual, 26 images are taken in two separate sessions. There are 13 images for each session, in which 3 images with sunglasses, another 3 with scarves and the remaining 7 are with different illumination and expression changes, the 13 images of one subject from Session 1 is shown in Fig. \ref{fig:AR_img}. Each image is 165$\times$120 pixels. In our experiments, we choose a subset of the AR database consisting of 50 men and 50 women, and all images are resized to 42$\times$30 pixels. The neutral face image of each subject is used as training data, and the sunglasses/scarf occluded images in each session for testing. The proposed method is compared with other state-of-the-art approaches, including HQPAMI \cite{he2013half}, NR \cite{qian2015robust}, ProCRC \cite{cai2016probabilistic}, F-LR-IRNNLS \cite{iliadis2017robust}, EGSNR\cite{zhang2020enhanced}, LDMR\cite{zhang2020locality}, and GD-HASLR \cite{wu2018occluded}. To better illustrate the superiority of CSOIGO, we also present the results of IGO-PCA-NNC\cite{tzimiropoulos2012subspace}, IGO-PCA-CRC and SOIGO-PCA-NNC. Table \ref{table:ARoneimage} summarizes the experimental results, one can see that CSOIGO achieves the highest recognition accuracy under all cases except for the sunglasses scenario of session 1. It has the best overall result, and the overall accuracy gain of CSOIGO over GD-HASLR and IGO-PCA-CRC is 4.5\% and 2.67\%, respectively. The above experimental results indicate our proposed CSOIGO is robust to real disguise even when a single training sample per person is available.

Next, we utilize two neutral face images per subject from Sessions 1 and 2 for training, and the test sets are identical with the first experiment. The results are reported in Table \ref{table:ARtwoimages}. As can be seen from Table \ref{table:ARtwoimages}, CSOIGO yields the best overall recognition accuracy, and it outperforms GD-HASLR by 2.92\%.
\begin{table}[]
\centering
\caption{Recognition accuracy (\%) of competing approaches on the AR database when only one neutral face image per subject from Session 1 is used as training sample, the dimension that leads to the best result for IGO and SOIGO based approaches is given in parentheses.}
\label{table:ARoneimage}
\begin{tabular}{cccccc}
\hline
Methods  & \multicolumn{2}{c}{Sunglasses} & \multicolumn{2}{c}{Scarf} & Overall \\ \hline
         & Session1      & Session2     & Session1    & Session2    &         \\
HQPAMI\cite{he2013half}        &   56.67            &    38.00          & 38.00             &     22.33        &  3875       \\         
NR\cite{qian2015robust}        &    28.33           &     16.67         &     29.67        &     17.33        &   23.00      \\
ProCRC\cite{cai2016probabilistic}        &    53.07           &     31.00         & 18.67            & 7.33            &  27.52       \\
F-LR-IRNNLS\cite{iliadis2017robust}       &      88.67         &     60.33         & 67.00             &   49.67          &   66.42      \\
EGSNR\cite{zhang2020enhanced}  &   84.00  &  54.00  &  70.33  &  48.33  &    64.16     \\
LDMR\cite{zhang2020locality}       &  68.33  &  45.67  &  59.67  &  34.00  &    51.92     \\
GD-HASLR\cite{wu2018occluded} & 92.00        & 66.67       & 82.67      & 58.67      & 75.00  \\ 
IGO-PCA-NNC\cite{tzimiropoulos2012subspace} & 89.00 (99) & 69.00 (99) & 73.33 (97) & 53.33 (96)& 71.17\\
IGO-PCA-CRC & \textbf{93.00 (85)} & 74.33 (92) & 81.67 (88) & 58.33 (95) & 76.83 \\
SOIGO-PCA-NNC & 88.67 (92) & 73.33 (96) & 80.33 (99) & 61.00 (88) & 75.83 \\
CSOIGO & 92.67 (89) & \textbf{76.67 (93)} & \textbf{83.33 (75)} & \textbf{65.33 (99)} & \textbf{79.50} \\ \hline
\end{tabular}
\end{table}

\begin{table}[]
\centering
\caption{Recognition accuracy (\%) of competing approaches on the AR database when two neutral face images (from Sessions 1 and 2) per subject are used as training samples, the dimension that leads to the best result for IGO and SOIGO based approaches is given in parentheses.}
\label{table:ARtwoimages}
\begin{tabular}{cccccc}
\hline
Methods  & \multicolumn{2}{c}{Sunglasses} & \multicolumn{2}{c}{Scarf} & Overall \\ \hline
         & Session1      & Session2     & Session1    & Session2    &         \\
HQPAMI\cite{he2013half}        &      61.33         &      59.33       & 44.67 &            48.00 &   53.33      \\
NR\cite{qian2015robust}        &     34.00          &      33.33       &     33.00        & 35.67    &     34.00    \\
ProCRC\cite{cai2016probabilistic}        &     53.00         &     54.67        &    18.00         & 17.67  &    35.84    \\
F-LR-IRNNLS\cite{iliadis2017robust}       &       90.33         &  87.67            & 78.67  &      76.00       &   83.17     \\
EGSNR\cite{zhang2020enhanced}    &   88.00  &  89.33  &  80.00 &  73.00  &    82.58     \\
LDMR\cite{zhang2020locality}      &   71.00  &  63.67  &  64.00 &  61.00  &    64.92     \\
GD-HASLR\cite{wu2018occluded} &   93.00     &    93.33    &   82.67    &   84.00    &  88.25 \\ 
IGO-PCA-NNC\cite{tzimiropoulos2012subspace} & 93.00 (182) & 91.67 (191) & 78.00 (199) & 74.00 (193) & 84.17 \\
IGO-PCA-CRC & 96.00 (128)  & 95.33 (116) & 85.00 (190)  & 84.00 (160) & 90.08  \\
SOIGO-PCA-NNC & 96.33 (187) & 92.67 (197)  & \textbf{86.33 (166)}  & 83.67 (189) & 89.75 \\
CSOIGO & \textbf{97.33 (144)} & \textbf{95.67 (124)} & 86.00 (119)  & \textbf{85.67 (198)} & \textbf{91.17} \\ \hline
\end{tabular}
\end{table}

\subsection{Comparison with CNN-based approaches}
In this subsection, we compare our proposed method with prevailing deep learning-based approaches. The first one is VGGFace \cite{parkhi2015deep} which is based on the VGGNet \cite{simonyan2014very}, and it has 16 convolutional layers, five max-pooling layers, three fully-connected layers and a final linear layer with softmax layer. In our experiments, we employ FC6 and FC7 for feature extraction. The second one is Lightened CNN \cite{wu2015lightened} which has a low computational complexity. Lightened CNN consists of two different models, \ie, Model A and Model B. Model A is based on the AlexNet \cite{krizhevsky2012imagenet}, which contains four convolution layers using the max feature map (MFM) activation functions, four max-pooling layers, two fully-connected layers, and a linear layer with softmax activation in the output. Model B is based on the Network in Network model \cite{lin2013network} and consists of five convolution layers using the MFM activation functions, four convolutional layers for dimensionality reduction, five max-pooling layers, two fully-connected layers, and a linear layer with softmax activation in the output. For Lightened CNN, FC1 is used for feature extraction. All the features extracted by VGGFace and Lightened CNN are classified using the nearest neighbor classifier with cosine distance. Like in subsection \ref{sec_41}, the first experiment is one neutral face of each subject for training on the AR database, and the experimental results are summarized in Table \ref{table:CNNoneImage}. Table \ref{table:CNNtwoImages} lists the results when two neutral faces are used for training. From Tables \ref{table:CNNoneImage} and \ref{table:CNNtwoImages}, we can see that VGGFace and Lightened CNN perform better in the scarf scenario than in the sunglasses scenario. This indicates that they have difficulty to tackle the upper face occlusion, and this phenomenon is also observed in Ref. \citenum{mehdipour2016comprehensive}. For Lightened CNN, Model A outperforms Model B. Whether one or two neutral face images per subject are used for training, our proposed CSOIGO achieves the best overall recognition accuracy.
\begin{table}[]
\centering
\caption{Comparison with CNN-based approaches on the AR database when only one neutral face image per subject from Session 1 is used as training samples, the dimension that leads to the best result for IGO and SOIGO based approaches is given in parentheses.}
\label{table:CNNoneImage}
\begin{tabular}{cccccc}
\hline
Methods  & \multicolumn{2}{c}{Sunglasses} & \multicolumn{2}{c}{Scarf} & Overall \\ \hline
         & Session1      & Session2     & Session1    & Session2    &         \\
VGGFace FC6\cite{parkhi2015deep}       &       54.00       &    45.00         & \textbf{91.67} &           \textbf{88.00} &   69.67     \\
VGGFace FC7\cite{parkhi2015deep}        &      45.67        &     40.00        & 88.67 &          84.00 &  64.59      \\
Lightened CNN (A)\cite{wu2015lightened}        &    67.33          &     56.00        &    87.00 &  82.33  & 73.17    \\
Lightened CNN (B)\cite{wu2015lightened}       &      36.33          &      31.33        &  80.67 &     73.67        &   55.50     \\
GD-HASLR\cite{wu2018occluded} & 92.00        & 66.67       & 82.67      & 58.67      & 75.00  \\ 
IGO-PCA-NNC\cite{tzimiropoulos2012subspace} & 89.00 (99) & 69.00 (99) & 73.33 (97) & 53.33 (96)& 71.17\\
IGO-PCA-CRC & \textbf{93.00 (85)} & 74.33 (92) & 81.67 (88) & 58.33 (95) & 76.83 \\
SOIGO-PCA-NNC & 88.67 (92) & 73.33 (96) & 80.33 (99) & 61.00 (88) & 75.83 \\
CSOIGO & 92.67 (89) & \textbf{76.67 (93)} & 83.33 (75) & 65.33 (99) & \textbf{79.50} \\ \hline
\end{tabular}
\end{table}

\begin{table}[]
\centering
\caption{Comparison with CNN-based approaches on the AR database when two neutral face images (from Sessions 1 and 2) per subject are used as training samples, the dimension that leads to the best result for IGO and SOIGO based approaches is given in parentheses.}
\label{table:CNNtwoImages}
\begin{tabular}{cccccc}
\hline
Methods  & \multicolumn{2}{c}{Sunglasses} & \multicolumn{2}{c}{Scarf} & Overall \\ \hline
         & Session1      & Session2     & Session1    & Session2    &         \\
VGGFace FC6\cite{parkhi2015deep}       &     44.67         &     51.00        & \textbf{91.67} &           \textbf{93.33} &    70.17   \\
VGGFace FC7\cite{parkhi2015deep}        &      41.67        &   44.67          & 88.67 &          89.33 &   66.08     \\
Lightened CNN (A)\cite{wu2015lightened}        &     64.67        &   58.33          &    86.67 & 85.33   &  73.75   \\
Lightened CNN (B)\cite{wu2015lightened}       &     38.67           &     38.00         &  81.67 &      79.33       &   59.42     \\
GD-HASLR\cite{wu2018occluded} &   93.00     &    93.33    &   82.67    &   84.00    &  88.25 \\ 
IGO-PCA-NNC\cite{tzimiropoulos2012subspace} & 93.00 (182) & 91.67 (191) & 78.00 (199) & 74.00 (193) & 84.17 \\
IGO-PCA-CRC & 96.00 (128)  & 95.33 (116) & 85.00 (190)  & 84.00 (160) & 90.08  \\
SOIGO-PCA-NNC & 96.33 (187) & 92.67 (197)  & 86.33 (166)  & 83.67 (189) & 89.75 \\
CSOIGO & \textbf{97.33 (144)} & \textbf{95.67 (124)} & 86.00 (119)  & 85.67 (198) & \textbf{91.17} \\ \hline
\end{tabular}
\end{table}

\subsection{Random block occlusion}
\label{sec_43}
\begin{figure}[t]
\centering
\includegraphics[trim={0mm 0mm 0mm 0mm},clip, width = .8\textwidth]{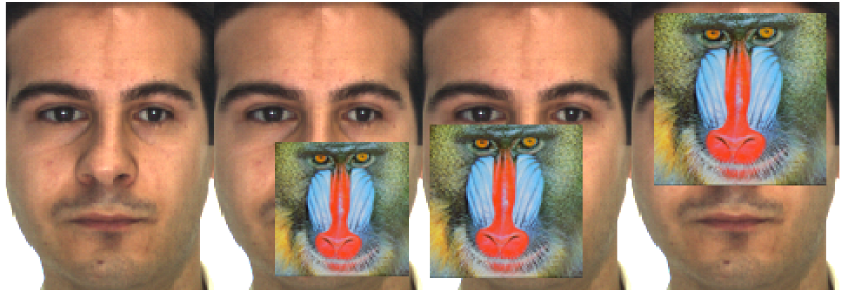}
\caption{Original face image and its occluded images with different occlusion percentages, from the second to the last the percentage is 30\%, 40\% and 50\%, respectively.}
\label{fig:occluded_face}
\end{figure}

Here, we conduct other experiments using synthesized occluded face data as testing data. For each subject, seven non-occluded face images in the AR dataset in Session 1 are used for training and the other seven non-occluded images in Session 2 for testing, the image size is 42$\times$30 pixels. Block occlusion is tested by placing the square baboon image on each test image. The location of the occlusion is randomly chosen and is unknown during training. We consider different sizes of the object such that the face is covered with the occluded object from 30\% to 50\% of its area, some occluded face images are shown in Fig. \ref{fig:occluded_face}. The above experimental results indicate that GD-HASLR is superior to other competing approaches; therefore, in this subsection and the following subsection, we report the result of GD-HASLR for comparison. Recognition results for different levels of occlusion are shown in Table \ref{table:occlusion}. One can see that CSOIGO outperforms GD-HASLR by a large margin, and the performance gain is significant with the increasing percentage of occlusion. Moreover, SOIGO-PCA-NNC outperforms IGO-PCA-NNC and CSOIGO performs better than IGO-PCA-CRC, which demonstrates that SOIGO is more robust than IGO when dealing with artificial occlusion.

\begin{figure}[t]
\centering
\includegraphics[trim={0mm 0mm 0mm 0mm},clip, width = .8\textwidth]{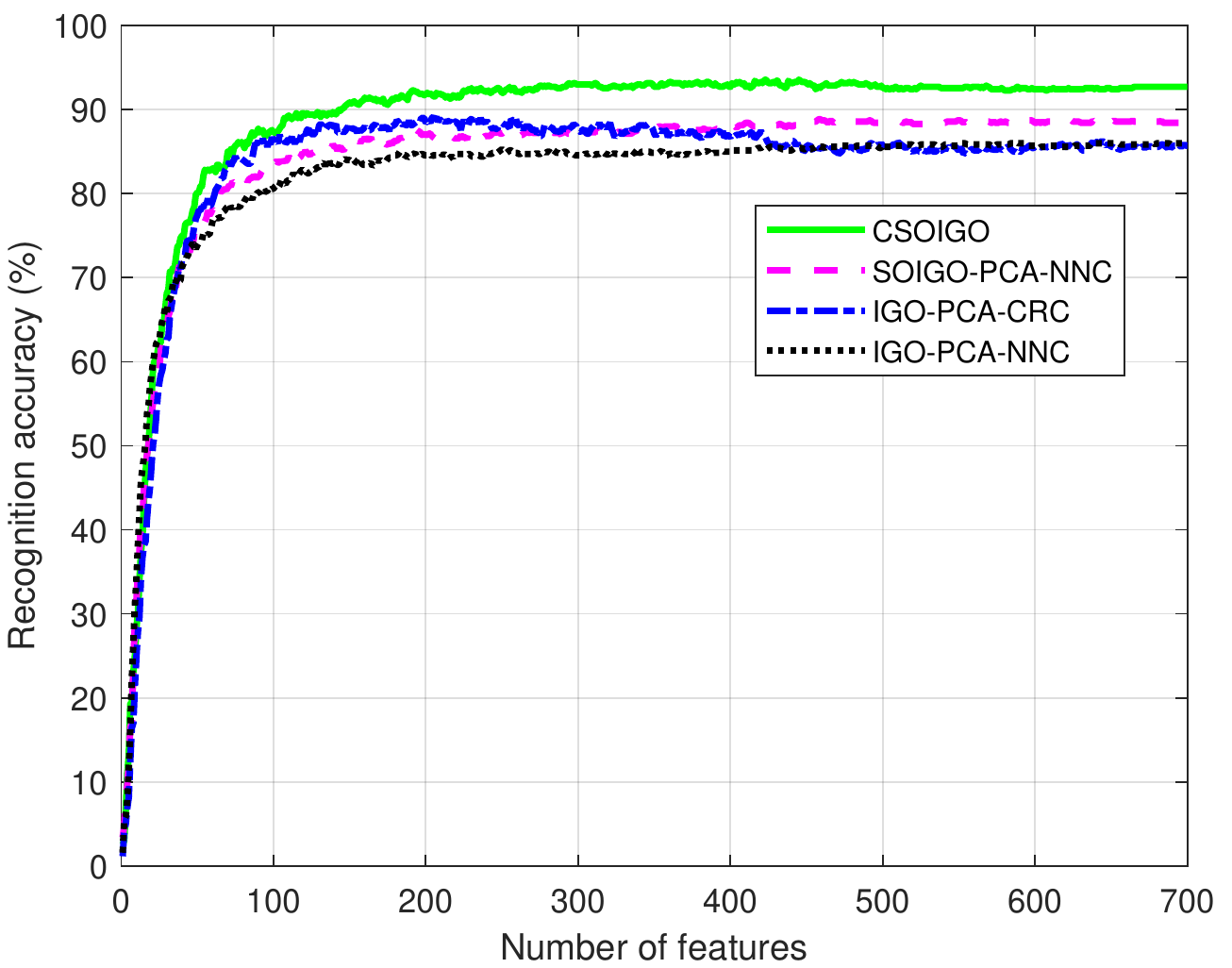}
\caption{Recognition accuracy versus different number of features when the percentage of occlusion is 30\%.}
\label{fig:percent30}
\end{figure}

To vividly show the performance of IGO and SOIGO based approaches under different number of features, in Fig. \ref{fig:percent30} we plot the recognition accuracy against the number of features when the percentage of occlusion is 30\%. We can clearly see that with the increasing of the number of features, CSOIGO consistently outperforms the other three competing approaches.

\begin{table}[]
\centering
\caption{Recognition accuracy (\%) of competing methods under different percentages of occlusion on the AR database, the dimension that leads to the best result for IGO and SOIGO based approaches is given in parentheses.}
\label{table:occlusion}
\begin{tabular}{cccc}
\hline
Occlusion percentage & 30\% & 40\% & 50\% \\ \hline
GD-HASLR\cite{wu2018occluded}             &  81.29    &  71.14    &  56.14    \\
IGO-PCA-NNC\cite{tzimiropoulos2012subspace}          &   86.14 (588)   &  80.57 (606)    &   66.29 (321)   \\
IGO-PCA-CRC          &  89.14 (205)    &  80.14(185)    &  71.29 (569)    \\
SOIGO-PCA-NNC        &  88.86 (458)    &  84.57 (575)    &  73.29 (693)    \\
CSOIGO        &  \textbf{93.57 (423)}    &   \textbf{87.00 (533)}   &   \textbf{76.57 (698)}   \\ \hline
\end{tabular}
\end{table}

\subsection{Recognition with mixed variations}
In this subsection, we evaluate our proposed CSOIGO and other compared approaches under the mixed variations.  As shown in Figs. \ref{fig:AR_img} (a) and (b), the first seven images per subject in Session 1 have variations of expression and illumination, thus
seven unoccluded images from Session 1 of the AR database are selected for training and another seven undisguised images from Session 2 are used for testing. Experimental results of compared methods are shown in Table \ref{table:acc_mixed}. We can see that CSOIGO has the best performance. Specifically, it makes 1.86\% and 0.86\% improvement over GD-HASLR and IGO-PCA-CRC, respectively.

\begin{table}[]
\centering
\caption{Recognition accuracy (\%) of compared approaches with mixed variations on the AR database, the dimension that leads to the best result for IGO and SOIGO based approaches is given in parentheses.}
\label{table:acc_mixed}
\begin{tabular}{cc}
\hline
Methods       & Accuracy (\%) \\ \hline
GD-HASLR\cite{wu2018occluded}      &   96.71            \\
IGO-PCA-NNC\cite{tzimiropoulos2012subspace}   &   93.14(478)            \\
IGO-PCA-CRC   &   97.71(100)            \\
SOIGO-PCA-NNC &   94.71(371)            \\
CSOIGO &   \textbf{98.57(171)}            \\ \hline
\end{tabular}
\end{table}

\section{Conclusion}
\label{sec_5}
In this paper, we present a new method for occluded face recognition, namely CSOIGO, by exploiting the second order gradient information. SOIGO is robust to real disguise, synthesized occlusion and mixed variations. By employing CRC as the final classifier, our proposed method achieves impressive results in various scenarios and even outperforms some deep neural network based approaches.

In future work, we will introduce SOIGO into other popular subspace learning approaches, \eg, linear discriminant analysis (LDA), to extract more discriminative features. Moreover, other variants of CRC will also be investigated to further enhance the performance of recognition.

\acknowledgments 
This work was supported in part by the National Natural Science Foundation of China (Grant 62020106012, Grant U1836218, Grant 61902153, Grant 61876072, Grant 62006097, Grant 61672265), in part by the Fundamental Research Funds for the Central Universities (Grant JUSRP121104), in part by the Natural Science Foundation of Jiangsu Province (Grant BK20200593), and the 111 Project of Ministry of Education of China (Grant B12018).


\bibliography{report}   
\bibliographystyle{spiejour}   


\vspace{1ex}\noindent\textbf{He-Feng Yin} received the B.S. degree in the School of Computer Science and Technology from Xuchang University, Xuchang, China, in 2011 and the Ph.D. degree from the School of Internet of Things Engineering, Jiangnan University, Wuxi, China, in 2020. Currently, he is a postdoctoral researcher in the School of Artificial Intelligence and Computer Science, Jiangnan University, Wuxi, China. He was a visiting PhD student at Centre for Vision, Speech and Signal Processing (CVSSP), University of Surrey, under the supervision of Prof. Josef Kittler. His research interests include representation based classification methods, dictionary learning and low rank representation.

\vspace{1ex}\noindent\textbf{Xiao-Jun Wu} received the B.Sc. degree in mathematics from Nanjing Normal University, Nanjing, China, in 1991, and the M.S. and Ph.D. degrees in pattern recognition and intelligent systems from the Nanjing University of Science and Technology, Nanjing, in 1996 and 2002, respectively. From 1996 to 2006, he taught at the School of Electronics and Information, Jiangsu University of Science and Technology, where he was promoted to Professor.

He has been with the School of AI \& CS, Jiangnan University, since 2006, where he is a Professor of Computer Science and Technology. He was a Visiting Researcher with the CVSSP, University of Surrey, U.K., from 2003 to 2004. He has published over 300 research papers in refereed international journals and conferences. He is an Associate Editor of Pattern Recognition Letters, International Journal of Computer Mathematics, and several other journals. His current research interests include pattern recognition and computational intelligence. He was a Fellow of the International Institute for Software Technology, United Nations University, from 1999 to 2000. He was a recipient of the Most Outstanding Postgraduate Award from the Nanjing University of Science and Technology.

\vspace{1ex}\noindent\textbf{Xiao-Ning Song} received the B.S. degree in computer science from Southeast University, Nanjing, China, in 1997, the M.S. degree in computer science from the Jiangsu University of Science and Technology, Zhenjiang, China, in 2005, and the Ph.D. degree in pattern recognition and intelligence system from the Nanjing University of Science and Technology, Nanjing, in 2010. He was a Visiting Researcher with the Centre for Vision, Speech, and Signal Processing, University of Surrey, Guildford, U.K., from 2014 to 2015. He is currently a Professor with the School of Artificial Intelligence and Computer Science, Jiangnan University, Wuxi, China. His current research interests include pattern recognition, machine learning, and computer vision.

\listoffigures
\listoftables

\end{spacing}
\end{document}